\documentclass[11pt,a4paper]{article}
\usepackage[hyperref]{acl2020}
\usepackage{times}
\usepackage{latexsym}

\usepackage{graphicx}
\usepackage{booktabs}
\usepackage{eqparbox}
\usepackage{arydshln}
\usepackage{multirow}
\usepackage{amsmath}
\usepackage{enumitem}

\newcommand{\fontmed}{\fontsize{9pt}{8.5pt}\selectfont}

\usepackage{microtype}

\aclfinalcopy 

\newcommand*\samethanks[1][\value{footnote}]{\footnotemark[#1]}

\title{How Additional Knowledge can Improve Natural Language Commonsense Question Answering?}
\def\aclpaperid{2404} 

\author{ Arindam Mitra \thanks{\quad These authors contributed equally to this work.} \:  Pratyay Banerjee\samethanks \: Kuntal Pal\samethanks \: Swaroop Mishra\samethanks \: Chitta Baral 
\\ Department of Computer Science, Arizona State University
\\ \texttt{amitra7,pbanerj6,kkpal,srmishr1@asu.edu,chitta}@asu.edu
}

\date{}

\begin{document}
\maketitle
\begin{abstract}
Recently several datasets have been proposed to encourage research in Question Answering domains where commonsense knowledge is expected to play an important role. Recent language models such as ROBERTA, BERT and GPT that have been pre-trained on Wikipedia articles and books have shown reasonable performance with little fine-tuning on several such Multiple Choice Question-Answering (MCQ) datasets. Our goal in this work is to develop methods to incorporate additional (commonsense) knowledge into language model-based approaches for better question-answering in such domains. In this work, we first categorize external knowledge sources, and show performance does improve on using such sources. We then explore three different strategies for knowledge incorporation and four different models for question-answering using external commonsense knowledge. We analyze our predictions to explore the scope of further improvements.

\end{abstract}

\section{Introduction}

\begin{figure}[t]
\begin{center}
   \includegraphics[width=0.8\linewidth]{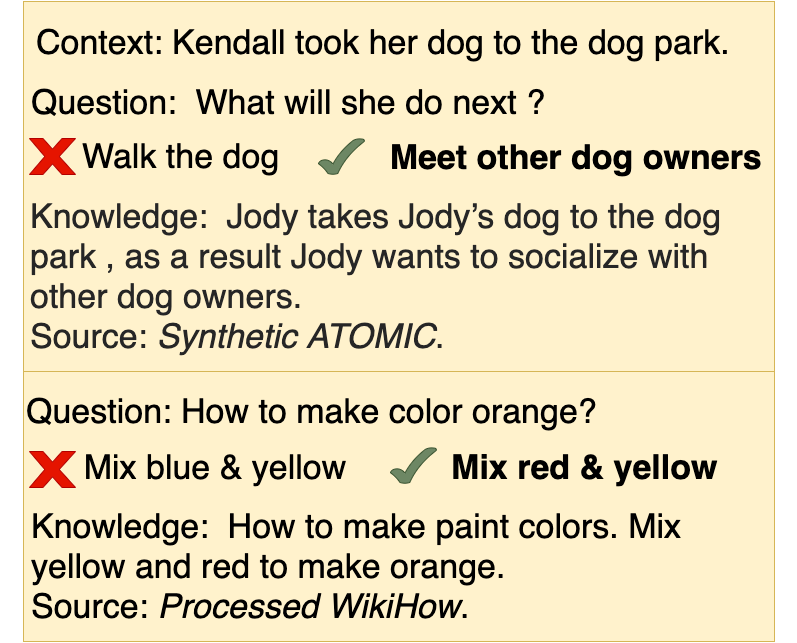}
\end{center}
   \caption{Above we have two strategies of knowledge incorporation. Below are sample questions from SocialIQA and PIQA, with corresponding top retrieved knowledge from our generated knowledge sources.}
\label{fig:motivation2}
\end{figure}

In recent months language models such as GPT \cite{radford2018improving}, BERT \cite{devlin-etal-2019-bert} and their variants (such as RoBERTa \cite{DBLP:journals/corr/abs-1907-11692}) that have been pre-trained on Wikipedia articles and books are able to perform very well on many of the natural language question-answering tasks. These days they form the defacto baseline for most new datasets. They even perform at near human level on many recently proposed natural language QA datasets \cite{rajpurkar2016squad,zellers2018swag}. These models do well even on some of the question-answering tasks where question-answering seemingly requires knowledge beyond what is given in the QA items. Perhaps it is because some of the needed knowledge that may be present in textual form is ``encapsulated'' by the language model-based systems as they are trained on a huge text corpus. But one may wonder 
whether more can be done; i.e., can the performance be improved by further infusion of the needed knowledge (or a knowledge base containing the needed knowledge), and what are ways of doing such knowledge infusion. DARPA and Allen AI upped the ante by developing several question-answering challenges where commonsense knowledge and reasoning are expected to play an important rule. The expected additional challenge in these domains is that often commonsense knowledge is not readily available in textual form \cite{Gordon:2013:RBK:2509558.2509563}. To answer the above-mentioned questions we consider three of those QA challenges: Abductive NLI (aNLI)\cite{bhagavatula2019abductive}, PIQA \cite{bisk2019piqa}, and Social Interaction QA (Social IQA) \cite{sap2019socialiqa}.

In this paper, we explore ways to infuse knowledge into any language model to reason and solve multiple-choice question-answering tasks. Considering a baseline performance of the BERT whole-word-masked model, we improve the performance on each of the datasets with three strategies. First, in the  \textit{revision strategy}, we fine-tune the BERT model on a knowledge-base (KB) which has knowledge statements relevant to that of each of the datasets and then use the model to answer questions. In the second, \textit{Open-Book Strategy}, we choose a certain number of knowledge statements from the KB that are textually similar to each of the samples of the datasets. Then we fine-tune the pre-trained BERT model for the question-answering task to choose the answer. In the final strategy, we take advantage of both the above-mentioned strategies. We first fine-tune the pre-trained BERT model on the KB and then use additional knowledge extracted for each sample for the question-answering.

To use the extracted knowledge from the KB, we propose four models, \textit{concat}, \textit{max}, \textit{simple sum}, and \textit{weighted sum}.
Each of the models uses knowledge in a different way to choose the correct answer among the options.

Apart from these we created a dataset, Parent and Family QA (PFQA) to analyze BERT's memorizing ability and to test BERT's ability for answering MCQ questions with necessary information scattered over multiple knowledge sentences.


Our contribution in this paper is as follows:
\begin{itemize}[noitemsep]
    \item We develop a framework for solving multiple-choice questions with external knowledge that can work on varied datasets.
    \item We propose four novel models representing four ways knowledge can be used with the language models.
    \item We study the three datasets, aNLI, PIQA, and Social IQA under three scenarios of external knowledge source respectively, and show that external knowledge helps. 
    In particular, we identify the models that takes us to be among the top in the leaderboard in the three datasets.
    \item We synthetically create a dataset, PFQA, and will make it publicly available. 
\end{itemize}

\section{Related Work}

    \paragraph{Question-Answering Datasets:}
    In datasets such as SQuAD \cite{rajpurkar2016squad}, TriviaQA \cite{joshi2017triviaqa}, 
    WikiQA \cite{yang2015wikiqa}, CoQA \cite{reddy2019coqa} the answers are present in either the passage or the context.
    Systems are able to achieve near-human performance on some of them. 
    HotpotQA \cite{yang2018hotpotqa} is a challenging dataset where the questions explicitly require multi-hop reasoning, and supporting knowledge passages derived from Wikipedia are provided.
   Another challenging QA task is when the multiple-choice questions do not have sufficient knowledge to answer correctly given a passage, context or options, as in ARC \cite{clark2018think}, RACE \cite{lai2017race}, and OpenBook QA \cite{OpenBookQA2018}. Recently language models trained on a huge corpus have been able to perform quite well \cite{devlin-etal-2019-bert,liu2019roberta} on them.
    Our focus in this paper is on datasets which not only require external facts but also need \textit{commonsense knowledge} to predict the correct answer, as in aNLI, PIQA  and Social IQA.

    \paragraph{External Knowledge:}
    Models that integrate external knowledge have been introduced in \cite{mihaylov2018knowledgeable,chen2018neural,yang2019enhancing,wang2019explicit}. These are aiming to utilize relational knowledge of the form $(a,R,b)$, where a and b are words, to modify pre-trained word vectors in both passages and questions to obtain better inter-word alignments. In our case, knowledge is more complex with $a$ and $b$ being event descriptions containing variables and thus computing alignment between knowledge passage and question-answer pair is more challenging.

    \paragraph{Knowledge Retrieval:}
    Systems for Information Retrieval, such as Elasticsearch \cite{gormley2015elasticsearch}, has been used in prior work of \cite{khot2019s,pirtoaca2019answering,Yadav2019QuickA,banerjee-etal-2019-careful,banerjee2019asu}. 
     In our work, we use elasticsearch for retrieval and we have a re-ranking algorithm using Spacy \cite{spacy2}. 
    
    

\begin{figure*}
\includegraphics[width=\linewidth]{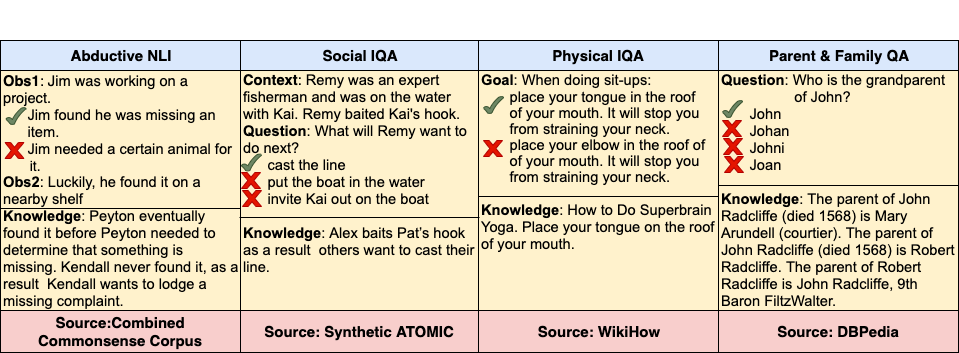}
   \caption{Example of all four datasets along with retrieved knowledge.}
\label{fig:motivation}
\end{figure*}

\section{MCQ Datasets}

  

In order to study how to incorporate knowledge, we need datasets which are designed to need external knowledge by question-answering systems. We choose four datasets to evaluate our models, each with a different kind of commonsense knowledge. Three of those are created by Allen AI researchers and one is generated synthetically by us.



\paragraph{Abductive NLI (aNLI):}
This dataset \cite{bhagavatula2019abductive} is intended to judge the potential of an AI system to do abductive reasoning in order to form possible explanations for a given set of observations. Given a pair of observations $O_1$ and $O_2$, the task is to find which of the hypothesis options $H_1$ or $H_2$ better explains the observations. There are 169,654 training and 1,532 validation samples. It also has a generation task, but we restrict ourselves to the multiple-choice task.

\paragraph{PIQA (Physical Interaction QA):}
This dataset is created to evaluate the physics reasoning capability of an AI system. The dataset requires reasoning about the use of physical objects and how we use them in our daily life. Given a goal $G$ and a pair of choices $C_1$ and $C_2$, the task is to predict the choice which is most relevant to the goal $G$. There are 16,113 training and 1,838 validation samples. 

\paragraph{SocialIQA:}
This dataset is a collection of instances about reasoning on social interaction and the social implications of their statements. Given a context $C$ of a social situation and a question $Q$ about the situation, the task is to choose the correct answer option $AO_i$ out of three choices. There are several question types in this dataset, which are derived from ATOMIC inference dimensions \cite{sap2019socialiqa,sap2019atomic}. In total, there are 33,410 training and 1,954 validation samples.

\paragraph{Parent and Family QA:}
We synthetically create this dataset to test both the memorizing capability of neural language models and the ability to combine knowledge spread over multiple sentences. The knowledge retrieved for the three earlier mentioned datasets may be error-prone and in some cases, absent. This is due to the errors from the Information Retrieval step. We create this synthetic dataset to have better control over the knowledge and to ensure that we have the appropriate knowledge to answer the questions.

The source of this dataset is DBPedia \cite{Auer:2007:DNW:1785162.1785216} from which we query for people and extract their parent information. Using this information, we generate 3 kinds of questions, which are, \textit{Who is the parent of X?}, \textit{Who is the grandparent of X?} and \textit{Who is the sibling of X?}. The dataset has a question $Q$ and 4 answer options $AO_i$. The names of a parent and their family members have many things in common, which can be used to answer such a question. To make the task harder, we remove the middle and last names from the answer options. To select wrong answer options, we select those names which are at an edit distance of one or two. We also evaluate the alternate strategy of using word-vector's cosine similarity to find the most similar names, but on evaluation we observe that the edit distance strategy creates confusing options.
For example, the most similar names to ``John'' are ``Robert'', and ``Williams'' using word-vectors. Using edit distance, we get ``Jon'', and ``Johan''. Since our models use word-piece tokenization, and corresponding embeddings, using edit distance makes the task harder.


We also ensure that all three kinds of questions for a particular person are present in the specific training and validation set. In total, there are 74,035 training, 9,256 validation and 9,254 test questions. 

\section{External Knowledge}
\subsection{Knowledge Categorization for Evaluation}
\label{sec:Kcat}
Reasoning with data from each of the above-mentioned datasets needs commonsense knowledge. We categorize external knowledge sources into three categories.

\paragraph{Directly Derived:} Here the commonsense QA task is directly derived from the knowledge source, and hence using the same knowledge may make the task trivial. We evaluate this on aNLI and the knowledge sources, \textit{ROCStories Corpus} \cite{mostafazadeh2016corpus} and \textit{Story Cloze Test} that were used in creating aNLI. Our motivation  is to see how well the model is able to answer questions when given the ``same'' knowledge.

\paragraph{Partially Derived:} Here the commonsense QA task is not directly derived from an external knowledge source, and considerable human knowledge was used to generate the question-answers. In this case, we use SocialIQA, which uses the \textit{ATOMIC}  \cite{sap2019atomic} knowledge base as the source for social events, but has undergone sufficient human intervention to make the task non-trivial. During dataset creation, the human turkers were asked to turn \textit{ATOMIC} events into sentences and were asked to create question-answers. 

\paragraph{Relevant:} Here the commonsense task is entirely created with the help of human turkers without use of a specific knowledge source.
But through our analysis of the question-answers, we
guess knowledge sources that seem relevant.
We evaluate this using PIQA as the commonsense task and WikiHow dataset \cite{koupaee2018wikihow} as the ``relevant'' external knowledge source.

\subsection{Knowledge Source Preparation}
\paragraph{aNLI:} For aNLI, we prepare multiple sets of knowledge sources. To test our first category of external knowledge, we use the entire \textit{Story Cloze Test} and \textit{ROCStories Corpus}. We also prepare another knowledge base that contains knowledge sentences retrieved for the train set of aNLI from the first knowledge base. This is done to not trivialize the task with knowledge leakage. We also create a knowledge source from multiple datasets such as \textit{MCTest} \cite{richardson-etal-2013-mctest}, \textit{COPA} \cite{roemmele2011choice} and \textit{ATOMIC}, but not \textit{Story Cloze Test} and \textit{ROCStories Corpus}. These sources contain commonsense knowledge which might be useful for the aNLI task. This Combined Commonsense Corpus belongs to the relevant knowledge category as described in section External Knowledge\ref{sec:Kcat}.

\paragraph{SocialIQA:} We synthetically generate a knowledge base from the events and inference dimensions provided by the \textit{ATOMIC} dataset \cite{sap2019atomic}. The \textit{ATOMIC} dataset contains events and eight types of if-then inferences \footnote{More details in Supplemental Materials.}. The total number of events is 732,723. Some events are masked, which we fill by using a BERT Large model and the Masked Language Modelling task \cite{devlin-etal-2019-bert}. We extend the knowledge source, and replace \textit{PersonX} and \textit{PersonY}, as present in the original ATOMIC dataset, using gender-neutral names. These steps may approximate the steps taken by humans to generate the question-answers.


\paragraph{PIQA:} We use the \textit{Wikihow} dataset for PIQA. It contains a large collection of paragraphs (214,544), each having detailed steps or actions to complete a task. We extract the title of each paragraph and split the paragraphs into sentences. The title is concatenated to each of the sentences. This is done to ensure the goal of the task should be present in each of the sentences.




\paragraph{Parent and Family QA:} We already possess the gold knowledge sentences. The knowledge for these questions is represented with simple sentences of the form ``The parent of X is Y''. We do not provide knowledge sentences for questions about \textit{grand parents} and \textit{siblings}. To answer such questions, the systems need to combine information spread over multiple sentences. Nearly all language models are trained over Wikipedia, so all language models would have seen this knowledge.

\begin{figure*}
  \includegraphics[width=\linewidth]{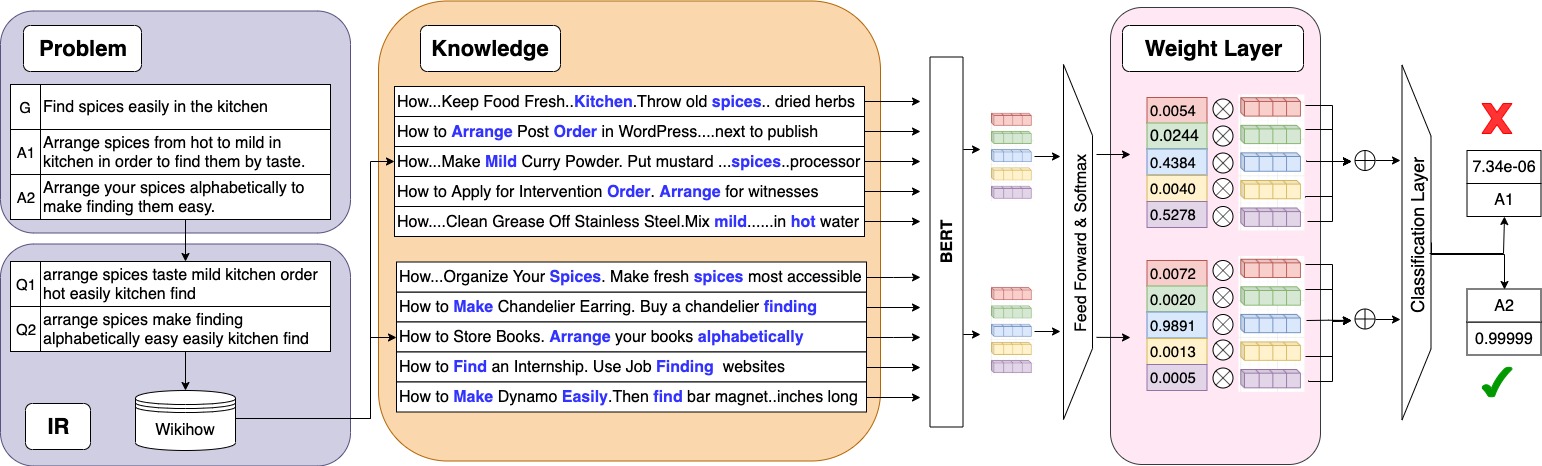}
  \caption{An end-to-end view of our approach. From query generation, knowledge retrieval, the different types of knowledge retrieved along with keywords highlighted in blue, the corresponding learned weights in the Weighted-Sum model and finally to predicted logits.}
  \label{fig:physeg1}
\end{figure*}

\subsection{Knowledge Retrieval}

\paragraph{Query Generation:} For query generation, we concatenate the question, answer option and the context if present, and remove standard English stopwords. We use verbs, adjectives, and adverbs from the question-answer pairs.

\paragraph{Information Retrieval System:} We use Elasticsearch to index all knowledge base sentences. We retrieve the top 50 sentences for each question-answer pair. The retrieved sentences may contain the key search words in any order.

\paragraph{Re-Ranking:} We perform Information Gain based Re-ranking using Spacy as described in \cite{banerjee-etal-2019-careful}. We use sentence similarity and knowledge redundancy to perform the iterative re-ranking. For similarity we use Spacy sentence similarity, for knowledge redundancy we find similarity with the already selected sentences. 
After re-ranking, we select the \textbf{top ten} sentences. 

We keep our Information Retrieval system generic as the tasks require varying kinds of commonsense knowledge; for example, If-then rules in SocialIQA, Scripts or Stories in aNLI, and understanding of Processes and Tools in PIQA.

\section{Standard BERT MCQ Model}
After extracting relevant knowledge from the respective KBs, we move onto the task of Question-Answering.
In all our experiments we use BERT's uncased whole-word-masked model ($BERT_{UWWM}$) \cite{devlin-etal-2019-bert} and RoBERTa \cite{liu2019roberta}.

\paragraph{Question-Answering Model:}
As a baseline model, we used pre-trained $BERT_{UWWM}$ for the question-answering task with an extra feed-forward layer for classification as a fine-tuning step.





\begin{table*}[t]
        \centering
        \scriptsize
        \resizebox{1.4\columnwidth} {!}{
        \begin{tabular}{c c c c c c}
            \toprule 
            \textbf{Dataset} & \multicolumn{1}{c}{\textbf{Strategy}} & \multicolumn{1}{c}{\textbf{Concat}} $\uparrow$  & \multicolumn{1}{c}{\textbf{Max}} $\uparrow$  & \multicolumn{1}{c}{\textbf{Sim-Sum}} $\uparrow$  & \multicolumn{1}{c}{\textbf{Wtd-Sum}} $\uparrow$ \\
            \hline
            \multirow{3}{*}{
                \rotatebox[origin=c]{0}{
                    \shortstack{\textbf{\texttt{aNLI}}}}}
            & O\textsc{nly}  O\textsc{penbook} & 73.89 & 73.69 & 73.50 & 73.26\\ 
            & O\textsc{nly} R\textsc{evision}  &72.65 & NA& NA& NA\\
            & R\textsc{evision} \& O\textsc{penbook} & 74.35 & 74.28 & 74.02 & \textbf{75.13} \\
             \hline
             \multirow{3}{*}{\rotatebox[origin=c]{0}{\textbf{\texttt{PIQA}}}}

            & O\textsc{nly}  O\textsc{penbook}  & 67.84 & 72.41 & 72.58 & 72.52\\
            & O\textsc{nly} R\textsc{evision}  & 74.53& NA& NA& NA\\
            & R\textsc{evision} \& O\textsc{penbook} & 67.74 & 73.83 & 76.76 & \textbf{76.82} \\
             \hline
             \multirow{3}{*}{\rotatebox[origin=c]{0}{\textbf{\texttt{SocialIQA}}}}
             & O\textsc{nly}  O\textsc{penbook} & 70.12 & 67.75 & 70.21 & \textbf{70.22}\\
            & O\textsc{nly} R\textsc{evision}  & 69.45& NA& NA& NA\\
            & R\textsc{evision} \& O\textsc{penbook} & 68.80 & 66.56 & 68.86 & 69.29 \\
             \hline
             \multirow{3}{*}{\rotatebox[origin=c]{0}{\textbf{\texttt{Parent \& Family QA}}}}
            &O\textsc{nly}  O\textsc{penbook} &91.21 &  89.8 & 92.66 & \textbf{92.8}6 \\
            & O\textsc{nly} R\textsc{evision} &78.30 & NA& NA& NA\\
            & R\textsc{evision} \& O\textsc{penbook} &87.21 &91.92 & 92.62 & {92.63} \\
             \hline
            \end{tabular}
            }
            \caption{
            Validation set accuracy (\%) of each of the four models (Concat, Max, Simple sum, Weighted sum) across four datasets for each of the three strategies. The base model is BERT Large whole-word-masked.
                }
            \label{tab:withk}
        \end{table*}   

\section{Modes of Knowledge Infusion}
We experiment with four different models of using knowledge with the standard BERT architecture for the open-book strategy. Each of these modules take as input a problem instance which contains a question $Q$, $n$ answer choices $a_1,...,a_n$ and a list called \textit{premises} of length $n$. Each element in \textit{premises} contains $m$ number of knowledge passages which might be useful while answering the question $Q$. Let $K_{ij}$ denotes the $j$-th knowledge passage for the $i$-th answer option. Each model computes a score $score(i)$ for each of the $n$ answer choices. The final answer is the answer choice that receives the maximum score. We now describe how the different models compute the scores differently.

\paragraph{Concat:}
In this model, all the $m$ knowledge passages for the $i$-th choice are joined together to make a single knowledge passage $K_i$. The sequence of tokens \{[CLS] $K_i$ [SEP] $Q a_i$ [SEP]\} is then passed to BERT to pool the [CLS] embedding from the last layer. This way we get $n$ [CLS] embeddings for $n$ answer choices, each of which is projected to a real number ($score(i)$) using a linear layer. 

\paragraph{Parallel-Max:}
For each answer choice $a_i$, Parallel-Max uses each of the knowledge passage $K_{ij}$ to create the sequence \{[CLS] $K_{ij}$ [SEP] $Q a_i$ [SEP]\} which is then passed to the BERT model to obtain the [CLS] embedding from the last layer which is then projected to a real number using a linear layer. $score(i)$ is then taken as the maximum of the $m$ scores obtained using each of the $m$ knowledge passage. 

\paragraph{Simple Sum:}
Unlike the previous model, \textit{simple sum} and the next two models assume that the information is scattered over multiple knowledge passages and try to aggregate that scattered information. To do this, the \textit{simple sum} model, for each answer choice $a_i$ and each of the knowledge passage $K_{ij}$ creates the sequence \{[CLS] $K_{ij}$ [SEP] $Q a_i$ [SEP]\} which it then passes to the BERT model to obtain the [CLS] embedding from the last layer. All of these $m$ vectors are then summed to find the summary vector, which is then projected to a scalar using a linear layer to obtain the $score(i)$.
\paragraph{Weighted Sum:}
The \textit{weighted sum} model computes a weighted sum of the [CLS] embeddings as some of the knowledge passage might be more useful than others. It computes the [CLS] embeddings in a similar way to that of the \textit{simple sum} model. It computes a scalar weight $w_{ij}$ for each of the $m$ [CLS] embedding using a linear projection layer which we will call as the \textit{weight layer}. The weights are then normalized through a softmax layer and used to compute the weighted sum of the [CLS] embeddings. It then uses (1) a new linear layer  or (2) reuses the weight layer (\textit{tied version}) to compute the final score $score(i)$ for the option $a_i$. We experiment with both of these options.

\begin{table}[t]
        \centering
        \scriptsize
        \resizebox{1.0\columnwidth}{!}{
        \begin{tabular}{c c c c}
            \toprule 
            \textbf{Dataset} & \multicolumn{1}{c}{\textbf{RoBERTA+Knowledge Source}} & \multicolumn{1}{c}{\textbf{Dev}} 
            \\
            \hline
            \multirow{3}{*}{
                \rotatebox[origin=c]{0}{
                    \shortstack{\textbf{\texttt{aNLI}}}}}
            & D\textsc{irectly} D\textsc{erived} & 86.68 \\ 
            & T\textsc{rain}O\textsc{nly} D\textsc{irectly} D\textsc{erived}  &   85.84\\
            & R\textsc{elated} K\textsc{nowledge} & 84.97   \\
             \hline
             
            \multirow{2}{*}{
                \rotatebox[origin=c]{0}{
                    \shortstack{\textbf{\texttt{SocialIQA}}}}}
            & P\textsc{artially} D\textsc{erived} &  79.53  \\ 
            & T\textsc{rain}O\textsc{nly} P\textsc{artially} D\textsc{erived}   & 78.85  \\
             \hline
             
            \end{tabular}
            }
            \caption{
            Comparison of Performance of aNLI and SocialIQA datasets across multiple knowledge sources for Weighted-Sum model. For ANLI, Directly Derived refers to ROC-Stories and Story Cloze datasets.  Related knowledge is Combined Commonsense Corpus. For SocialIQA, Partially Derived is the synthetic ATOMIC. For both, TrainOnly is the knowledge extracted only from the train set.
                }
            \label{tab:eval}
        \end{table}  

\begin{table}[t]
        \centering
        \tiny
        \begin{tabular}{c c c c}
            \toprule 
            \textbf{Dataset} & \multicolumn{1}{c}{\textbf{Model}} &
            \multicolumn{1}{c}{\textbf{Test}}\\
            \hline
            \multirow{2}{*}{
                \rotatebox[origin=c]{0}{
                    \shortstack{\textbf{\texttt{aNLI}}}}}
            
            & Baseline : B\textsc{ERT} 
            & 66.75 \\
            & Baseline : R\textsc{oBERTa}  
            & 83.91 \\
            & Our : B\textsc{ERT} 
            &74.96\\ 
            & Our : R\textsc{oBERTa}  
            & \textbf{84.18} \\
             \hline
             
            \multirow{2}{*}{
                \rotatebox[origin=c]{0}{
                    \shortstack{\textbf{\texttt{PIQA}}}}}
            & Baseline : B\textsc{ERT} 
            & 69.23\\
            & Baseline : R\textsc{oBERTa}  
            & \textbf{79.40} \\
            & Our :  B\textsc{ERT}
            & 72.28\\ 
            & Our :  R\textsc{oBERTa}  
            & 78.24 \\
             \hline
            \multirow{2}{*}{
                \rotatebox[origin=c]{0}{
                    \shortstack{\textbf{\texttt{SocialIQA}}}}}
            & Baseline : B\textsc{ERT} 
            & 64.50 \\
            & Baseline : R\textsc{oBERTa}  
            & 76.74 \\
            & Our :  B\textsc{ERT} 
            & 67.22\\ 
            & Our :  R\textsc{oBERTa}  
            & \textbf{78.00} \\
             \hline
             
            \multirow{2}{*}{
                \rotatebox[origin=c]{0}{
                    \shortstack{\textbf{\texttt{Parent \& Family QA}}}}}
             & Baseline : B\textsc{ERT} 
             & 76.96\\
            & Baseline : R\textsc{oBERTa}  
            & 78.36 \\
            & Our :  B\textsc{ERT}
            & 91.24\\ 
            & Our :  R\textsc{oBERTa} 
            & \textbf{93.40}\\
             \hline
            \end{tabular}
            \caption{
            Performance of the best knowledge infused model on the Test set. Best model scores are in bold.
                }
            \label{tab:eval2}
        \end{table}  

\section{Experiments}
Let $D$ be an MCQ dataset and $T$ be a pre-trained language model, $K_D$ be a knowledge base (a set of paragraphs or sentences) which is useful for $D$ and let $K$ be a general knowledge base where $T$ was pre-trained and $K$ might or might not contain $K_D$. We consider three approaches to infuse knowledge. 
\paragraph{Revision Strategy:}
In this strategy, T is fine-tuned on $K_D$ with respect to Masked LM and the next sentence prediction task and then fine-tuned on the dataset D with respect to the Question-Answering task.
\paragraph{Open Book strategy:}
Here a subset of $K_D$ is assigned to each of the training samples on the dataset D and the model $T$ is fine-tuned on the modified dataset $D$. 
\paragraph{Revision along with an Open Book Strategy:}
In this strategy, $T$ is fine-tuned on $K_D$ with respect to Masked LM and the next sentence prediction task and also a subset of $K_D$ is assigned to each of the training samples on $D$. The model is then fine-tuned with respect to the modified dataset as a Question-Answering task. 

\begin{figure}
  \centering
  \includegraphics[width=5.36cm]{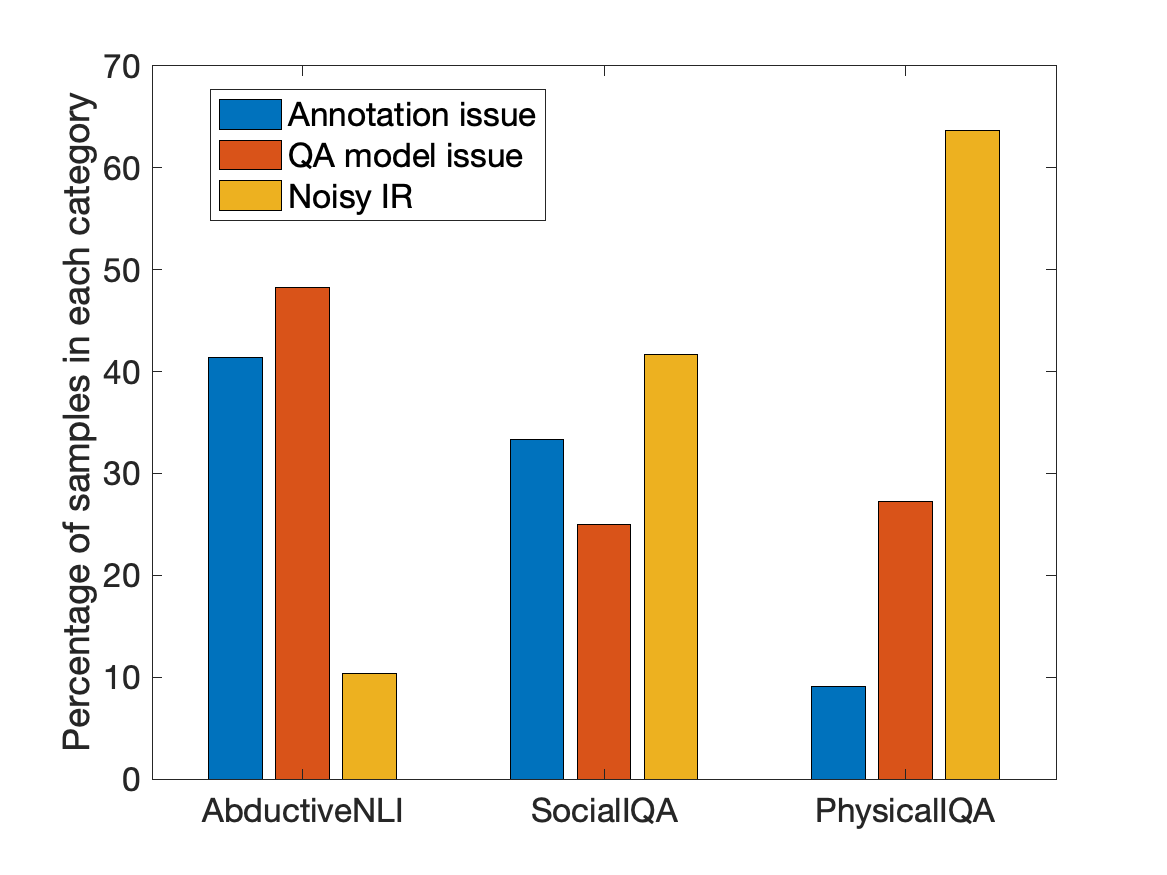}
  \caption{Different categories of errors.}
  \label{fig:wrongcs}
\end{figure}

\subsection{Results}
\noindent
\textbf{Which Strategy Works?} Table \ref{tab:withk} and Table \ref{tab:eval} summarize our experiments on four datasets. We observe that knowledge helps in improving the performance. Both the Open Book and the Revision strategy perform well. Together the performance improves even further. 
The performance of the Revision strategy is poor for the SocialIQA and PFQA datasets. The reason behind this drop in performance may be due to the synthetic nature of the sentences and the unavailability of next sentence prediction task data. Note that the knowledge in the KB for SocialIQA and PFQA are single sentences and not paragraphs. The results for PIQA and aNLI datasets are better due to the presence of natural and contiguous knowledge sentences. 

\begin{figure}
  \centering
  \includegraphics[width=5.36cm]{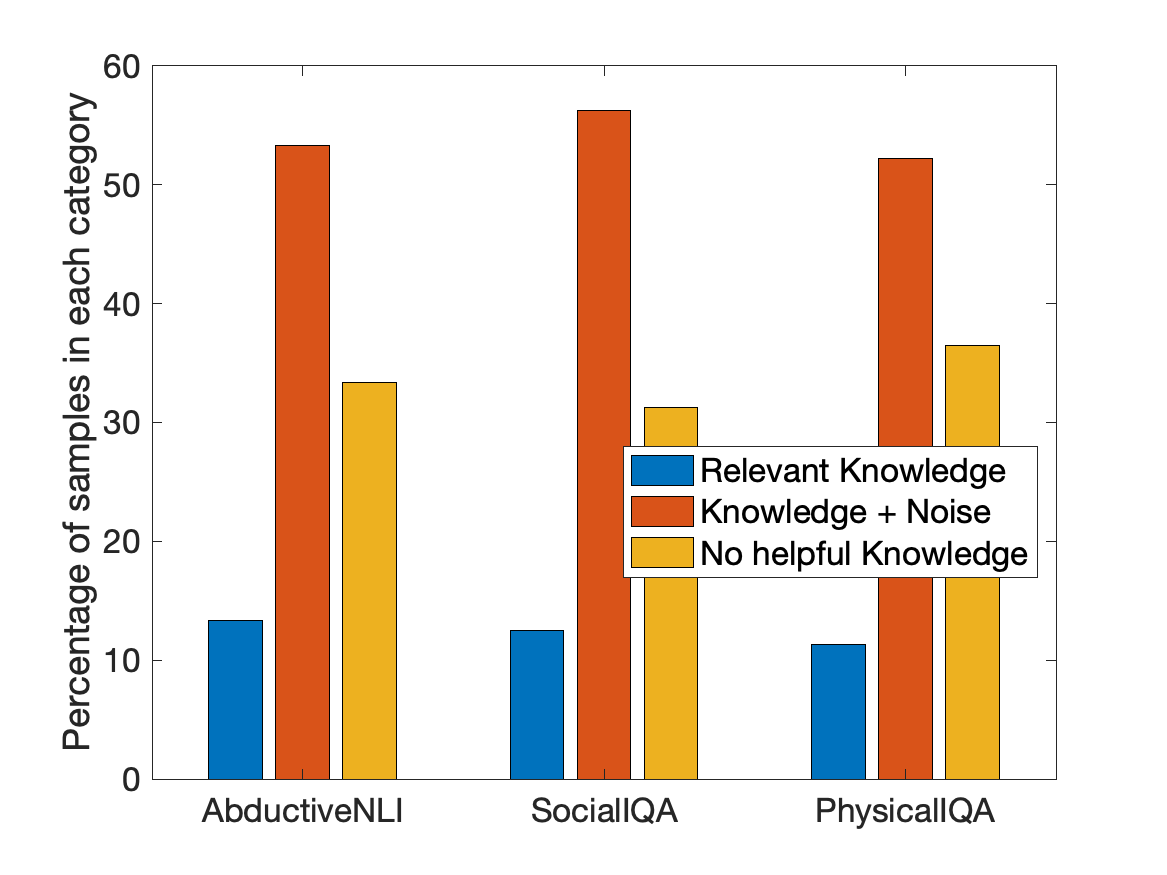}
  \caption{Categories of knowledge relevance for correct predictions.}
  \label{fig:correctcs}
\end{figure}

For PIQA, the BERT model improves with knowledge, whereas RoBERTa model underperforms, indicating RoBERTa gets distracted by the retrieved knowledge, and the knowledge it possess from pre-training is more useful. 

\noindent
\textbf{How did different External Knowledge Categories Perform ?} In Directly derived knowledge category, the model accuracy with knowledge is significantly more than the baseline accuracy. However, model is still not able to answer all questions because of two reasons. First, the model fails to reason well. Second, many hypotheses are possible between two observations, and turkers seem to have created hypotheses which are very different from the source data\footnote{More details in Supplemental Materials.}. In partially derived and relevant knowledge category, the model accuracy increases with addition of knowledge.

\begin{figure}
  \includegraphics[width=7cm]{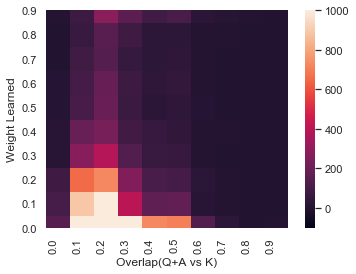}
  \caption{Weights learned by the RoBERTa Weighted-Sum Model vs the Normalized Overlap between knowledge and concatenated question-answer for all samples of PIQA validation set.}
  \label{fig:roberta_wt_overlap}
\end{figure}

\noindent
\textbf{Which is the best model for Knowledge Infusion?}
Weighted Sum model seems to be the best model for Knowledge Infusion. It is also partially explainable by looking at the weights associated with the knowledge sentences.

\begin{figure}
  \centering
  \includegraphics[width=5.5cm]{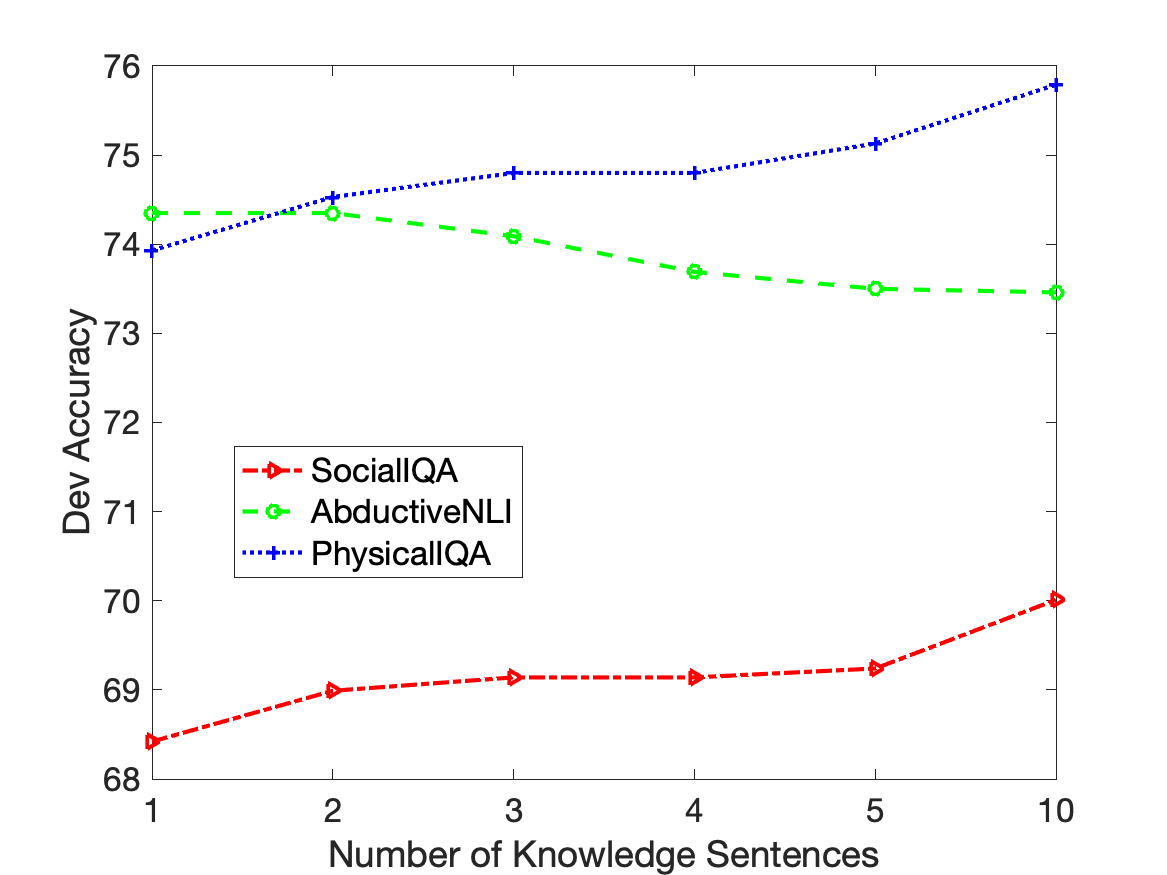}
  \caption{Validation accuracy versus number of retrieved knowledge sentences, for all three datasets.}
  \label{fig:varir}
\end{figure}

\noindent
\textbf{How many Knowledge Sentences do we need?} We experiment by varying number of Knowledge Sentences, and Figure \ref{fig:varir} illustrates that accuracy for SocialIQA and PIQA increases with more number of knowledge sentences, whereas accuracy decreases for aNLI. This is because, we use directly derived knowledge source for aNLI, so increasing the number of knowledge sentences acts as noise for aNLI. For SocialIQA and PIQA more knowledge helps.
\paragraph{Is there a correlation among overlap between question-answer \& knowledge and learned weights?}
Figure \ref{fig:roberta_wt_overlap} shows the weight versus overlap between knowledge and question-answer distribution for PIQA. It shows there is a low overlap, but the model learns to give high weights in some cases regardless of the overlap.

\section{Discussion and Error Analysis}
We have analyzed 200 samples from each of our best models, and the results are presented in Figures \ref{fig:correctcs} and \ref{fig:wrongcs}. Figure \ref{fig:correctcs} shows that around two third of correct predictions are because of the relevant knowledge provided in open book format. The figure also shows that many a times knowledge is acompanied with noise, and models are doing a good job in ignoring noise and attending to relevant knowledge. The Noisy IR category says model is also doing a good job in ignoring the complete open book knowledge in case its noisy. In those cases, either knowledge acquired during revision phase or the orignial language model training phase helps in answering correctly.
\\
\noindent
\textbf{Why did the weighted-sum model work best?} Weighted sum model provides the flexibility to attend in varying amount to multiple knowledge sentences, which is not true for rest of the models. \\
\noindent
\textbf{How to choose external knowledge?} External knowledge source needs to be chosen based on the domain to which the dataset belongs; e.g. there are lots of questions in PIQA which can be benefitted by answering how of various physical processes. Use of Wikihow dataset improves model accuracy.\\
\noindent
\textbf{How to choose a strategy?} Open book strategy should always be chosen. We saw in Figure \ref{fig:correctcs} that models are good at ignoring noise. This helps in handling those cases where knowledge retrived from IR is not relevant.
Revision strategy is also recommended unless sufficient data is not available for next sentence prediction task.\\
\noindent
\textbf{How do models perform in PFQA?} We observe neural language models are able to memorize and combine knowledge spread over multiple sentences. Most of the errors are observed in questions regarding grandparents and siblings, which indicate that there is still scope for improvement in multi-hop reasoning.\\
\noindent
\textbf{Use of RocStories for aNLI and ATOMIC for Social?} As discussed in section External Knowledge \ref{sec:Kcat}, in case of aNLI we wanted to check how well the model 
performs with the same knowledge that is used during the creation of the dataset. 
Similarily in case of SocialIQA we wanted to see how much is the accuracy boost if we provide partially derived knowledge. As explained earlier, for both cases, although the models do well, they do not reach near human accuracy.\\
\noindent
\textbf{What are the category of errors models made?}
We divide the errors into three categories. Category I is annotation issue, which is the case when more than one answer option is correct or incorrect answer option is labelled correct. Also, questions for which information is insufficient to select a specific answer option falls in this category. For example:\\
\\
\noindent
\fbox{\begin{minipage}{0.97\linewidth}
\fontmed
   \textbf{Obs1:} Alan got a gun for his 18th birthday.\\
   \textbf{Obs2:} He now loves to go hunting.
   \\(Hyp1) His dad took him hunting.\\  \textbf{(Hyp2) Alan decided to go hunting}
\end{minipage}}\\
\\Category II is where the IR output is noisy, and does not have relevant knowledge. For example:\\

\noindent
\fbox{\begin{minipage}{0.97\linewidth}
\fontmed
   \textbf{Question:} Blankets
   \\\textbf{(a) can cover lights}  (b) can cover candles \\\textbf{Knowledge: }How to Pack for Self Storage  .    Stand sofas on end to save space and cover your sofas with plastic covers and blankets.
\end{minipage}}\\
\\Category III is Question-Answering model issue  where relevant knowledge is present, though the knowledge is not completely exact\footnote{More details with BERT prediction analysis in Supplemental Materials.}. However, some reasoning with the help of this relevant knowledge could have helped the model in predicting the correct answer. For example:\\

\noindent
\fbox{\begin{minipage}{0.97\linewidth}
\fontmed
   \textbf{Obs1:} Tim needed a fruit to eat.
   \textbf{Obs2:} Finally, he found some fresh grapes to eat.
   \\(Hyp1) He went to the near by super market.  \textbf{(Hyp2) Tim looked for a long time in the messy fridge.}\\\textbf{Knowledge:}Tim needed a fruit to eat. He wanted a fruit that tasted good. He looked in the kitchen for the fruit. He almost gave up.  Finally  he found some fresh grapes to eat. 
\end{minipage}}\\


\noindent
\textbf{Future work}
In future we will work on better reasoning models and apply semantic IR to reduce QA Model issue and Noisy IR issue, which might lead to better performance. Some of the errors are because of annotation issue. Those might be excluded for calculating the exact accuracy.





\section{Conclusion}
Although this paper is about analyzing different ways to incorporate knowledge into language models in commonsense QA, we note that we are among the top of the leaderboard in the three tasks, SocialIQA, aNLI, and PIQA. We have provided four new models for multiple-choice natural language QA using the knowledge and analyzed their performance on these commonsense datasets. We also make a synthetic dataset available which measures the memorizing and reasoning ability of language models. We observe that existing knowledge bases even though do not contain all the knowledge that is needed to answer the questions, do provide a significant amount of knowledge. Language models utilize some of the knowledge; still, there are areas where the models can be further improved, particularly the ones where the knowledge is present but the model could not answer, and where it predicted wrong answers with irrelevant knowledge.


\clearpage


\bibliography{anthology,acl2020}
\bibliographystyle{acl_natbib}

\appendix

\clearpage

\section{Supplemental Material}
\label{sec:supplemental}
We have four sections here. In the first section, we have analyzed BERT prediction for all 3 datasets. The second section is regarding analysis of our dataset aNLI, where we have tried to find out why the model is not able to answer all questions even with the help of directly derived knowledge source. In the third section, we have compared ROBERTA and BERT by analyzing their weighted-sum model. Fourth section illustrates some examples in these datasets which need knowledge beyond common sense knowledge. 
\subsection{BERT prediction analysis}
To understand how knowledge is used in BERT and whether the knowledge is useful or not, we do the following analysis:
For each of the datasets we have randomly selected 100 samples where our best performing model predicts correctly and 100 samples where it has failed. We identified the following broad categories of analysis.

\begin{figure}[ht]
\small
  \includegraphics[width=7cm]{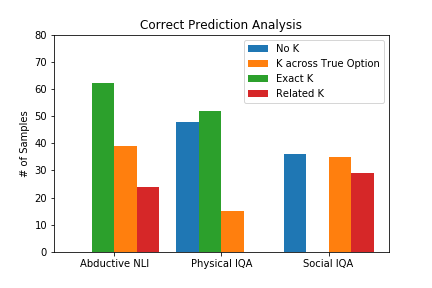}
  \caption{Measure of performance across different knowledge presence in correct predictions}
  \label{fig:corr}
\end{figure}

For the correct predictions, we check, (1) Exact appropriate knowledge is present, (2) A related but relevant knowledge is present, (3) Knowledge is present only in the correct option, and (4) No knowledge is present. Figure \ref{fig:corr} shows the counts for the above categories. All the cases do not occur in all the datasets. 

\begin{figure}[ht]
\small
  \includegraphics[width=7cm]{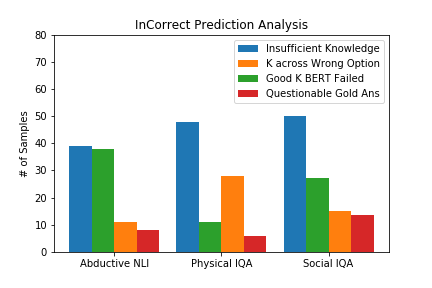}
  \caption{Measure of performance across different knowledge presence in incorrect predictions.}
  \label{fig:incorr}
\end{figure}

For the errors (Figure \ref{fig:incorr}), we analyze, (1) Is the knowledge insufficient, (2) Is the knowledge present in the wrong answer, (3) Knowledge is appropriate but model fails, and (4) Gold label is questionable. 

We also analyze given appropriate knowledge, how the model performs. From Figure \ref{fig:corr}, it can be seen that BERT can answer quite a number of question without knowledge. Also from Figure \ref{fig:incorr}, it is clear that despite of having good knowledge, BERT fails to answer correctly. 

In the following subsections, we analyze the different dataset specific errors.

\subsubsection{SocialIQA}
We measure the performance across the eight different ATOMIC inference dimensions for the best knowledge infused model. The six of the inferential dimensions are Needs, Attributes, Reactions, Wants, Motivations, Effects. These are for PersonX. There are two more for Others, Reaction and Wants. 

\begin{figure}[ht]
  \includegraphics[width=7.5cm]{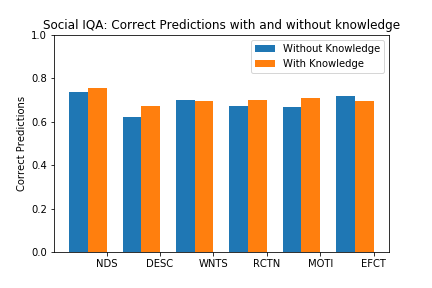}
  \caption{Performance of the model with (MAC model) and without knowledge (Baseline) across different types of ATOMIC inference dimensions.}
  \label{fig:soc}
\end{figure}

In figure \ref{fig:soc} we can see both with and without knowledge the model performs nearly equally across all dimensions. There is no considerable improvement across any particular dimension. 

In some cases the model fails to predict the correct answer despite of the appropriate knowledge being present. 

\fbox{\begin{minipage}{19em}
    \textbf{Question:} \textit{Kendall took their dog to the new dog park in the neighborhood. . What will Kendall want to do next?} \\ (A) \textit{walk the dog} (B) \textbf{meet other dog owners}
    \textbf{Knowledge:} Jody takes Jody's dog to the dog park, as a result  Jody wants to socialize with other dog owners.
\end{minipage}}
In the above example, the above knowledge was retrieved but still the model predicted the wrong option. 341 questions were predicted wrongly after addition of knowledge.
We also identified out of the set of 100 analyzed correct predictions, 29\% of the questions had partial information relevant to the question.

\begin{figure}[ht]
  \includegraphics[width=7.5cm]{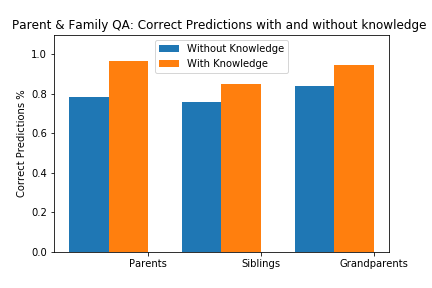}
  \caption{Performance of the model across the three different type of questions.}
  \label{fig:paf}
\end{figure}

\subsubsection{Parent and Family QA}
In Figure \ref{fig:paf}, we see with addition of knowledge, there is a considerable improvement in performance. Other than questions asking about parents, which just need a look up to answer, the sibling and grandparent questions need models to combine information present across multiple sentences. We can see the model improves even in this questions, showing knowledge infusion helps. Out of the three types of the questions, the performance is lowest on the sibling questions, indicating that it is harder for the models to perform this task. 
The model accuracy is reasonably good on this dataset, which shows BERT has a strong capability to memorize factual knowledge. Its performance improves with infusion of knowledge, 

Here also, 1,790 questions which were previously predicted correctly, are predicted wrong with addition of knowledge.

\subsubsection{PIQA}

Out of the 100 failures that we have analysed, we found that for 8 samples the \textit{goal} matches the knowledge statements but the answers present in the knowledge is different.
For example,

\fbox{\begin{minipage}{19em}
    \textbf{Goal:} \textit{How can I soothe my tongue if I burn it?} \\ (A) Put some salt on it. \textbf{(B) Put some sugar on it.}\\
    
    \textbf{Knowledge:} How to Soothe a Burnt Tongue.Chew a menthol chewing gum.
\end{minipage}}

Also, there are 33 samples in the whole train and dev dataset for which the words in one options are a subset of second option. In those cases, the knowledge retrieved is same for both the options and this confuses the BERT model.

\fbox{\begin{minipage}{19em}
    \textbf{Goal:} \textit{What can I drink wine out of if I don't have a wine glass?} \\ (A) Just pour the wine into a regular mug or glass and drink. \textbf{(B) Just pour the wine into a regular mug or wine glass and drink.}\\
    \textbf{Knowledge:} How to Serve Foie Gras. Pour a glass of wine.
\end{minipage}}

On addition of knowledge, 359 samples have become correctly predicted with our best model for PIQA dataset which were initially incorrect. But in the process, 166 samples which were correct in our baseline model have now been incorrectly predicted. 

\subsubsection{aNLI}
In this dataset, we also have some examples where negative knowledge is being fed to the model, and it still produces the correct output. There are 8 such examples among the 100 samples we analyzed. For example:

\fbox{\begin{minipage}{19em}
    \textbf{Obs1:} \textit{Pablo likes to eat worms.} \\
    \textbf{Obs2:} \textit{Pablo does not enjoy eating worms.} \\
     (Hyp1) Pablo thought that worms were a delicious source of protein. \textbf{(Hyp2) Pablo then learned what worms really are.}\\
     
     \textbf{Knowledge:} Pablo likes to eat worms. He read a book in school on how to do this. He fries them in olive oil. He likes to do this at least once a month. Pablo enjoys worms and views them as a delicacy.
\end{minipage}}

Similarily, we have examples where knowledge favors incorrect hypothesis, however our system still produces correct output. We found 12 such examples among the 100 samples we analyzed. For example:

\fbox{\begin{minipage}{19em}
    \textbf{Obs1:} \textit{Dotty was being very grumpy.} \\
    \textbf{Obs2:} \textit{She felt much better afterwards.} \\
    (Hyp1) Dotty ate something bad. \textbf{(Hyp2) Dotty call some close friends to chat.}\\
    \textbf{Knowledge:} Allie felt not so good last night. She ate too much. So she had to sleep it off. Then she woke up. She felt so much better
\end{minipage}}

We have 12 cases among 100 analyzed samples, where both hypothesis are very similar. So,our system is unable to produce correct output. For example:

\fbox{\begin{minipage}{19em}
    \textbf{Obs1:} \textit{Bob's parents grounded him.} \\
    \textbf{Obs2:} \textit{He came back home but his parents didn't even know he left.} \\
    (Hyp1) Bob got caught sneaking out. \textbf{(Hyp2) Bob got away with sneaking out.}\\
\end{minipage}}

We also have 34 examples where incorrect hypothesis has more word similarity with the observation and knowledge, whereas correct hypothesis has been paraphrased or has less word similarity. The system predicts the wrong answer in such a situation. One such example is:

\fbox{\begin{minipage}{19em}
    \textbf{Obs1:} \textit{Mary's mom came home with more bananas than they could possibly eat.} \\
    \textbf{Obs2:} \textit{That was the best way ever to eat a banana!} \\
    \textbf{(Hyp1) Mary and her mom decided to make chocolate covered frozen bananas to avoid waste.} (Hyp2) Mary made pineapple splits for everyone.\\
    \textbf{Knowledge:} Mary s mom came home with more bananas than they could possibly eat. She wondered why she had bought them all. Then after dinner that night she got a surprise. Mom made banana splits for the whole family. That was the best way ever to eat a banana
\end{minipage}}

Another area where the system fails, is where the problem seems to be open-ended, and many hypotheses can explain the pair of observations. It is tough to find exact knowledge in such a scenario. For example,

\fbox{\begin{minipage}{19em}
    \textbf{Obs1:} \textit{Lisa went for her routine bike ride.} \\
    \textbf{Obs2:} \textit{Some days turn out to be great adventures.} \\
    \textbf{(Hyp1) Lisa spotted a cat and followed it off trail} (Hyp2) Lisa saw a lot of great food.\\
    \textbf{Knowledge:} Lisa went for her routine bike ride.Only this time she noticed an abandoned house.She stopped to look in the house.It was full of amazing old antiques.Some days turn out to be great adventures.
\end{minipage}}
\subsection{Why Language model is not able to answer all questions even with the help of directly derived knowledge source?}

We experiment by providing directly derived knowledge source in aNLI, and find that the language model is still not able to answer all questions. 
We analyze and find following insights regarding aNLI and direcly derived knowledge source.
\begin{enumerate}
    \item 41.90 \% of the data has either Obs1 or Obs2 or both common with the knowledge.
    \item 23.12 \% of the data has both Obs1 and Obs2 common with the knowledge.
\end{enumerate}

\begin{figure}
  \includegraphics[width=\linewidth]{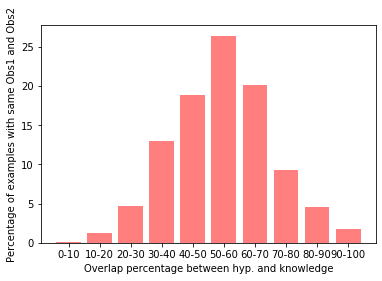}
  \caption{Overlap percentage between IRed knowledge and aNLI}
  \label{fig:overlap}
\end{figure}

Among those, figure \ref{fig:overlap} illustrates percentage overlap of hypothesis with knowledge. Overlap has been calculated by taking set of words in hypothsis, and finding if the same word is there in knowledge or not. You can see that very few examples have high overlap with knowledge. Actual number will be even lesser, as we have calculated word overlap, not phrase or sentence overlap.

\begin{figure}
  \includegraphics[width=\linewidth]{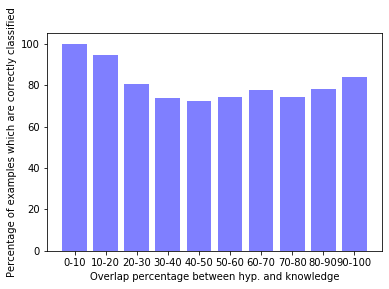}
  \caption{Percentage of samples correctly classified by best model in aNLI}
  \label{fig:score}
\end{figure}

For each of the bins, \ref{fig:score} illustrates how much of those are correctly classified. It is interesting to see that, even though knowledge is very much relevant here (obs1,obs2 same and high hyp overlap), BERT has not been able to classify correctly. You can ignore 0-10, 10-20 bins here, since percentage of data in those bins are very less as we saw in figure \ref{fig:overlap}.

\subsection{ROBERTA vs BERT}

\begin{figure}
  \includegraphics[width=\linewidth]{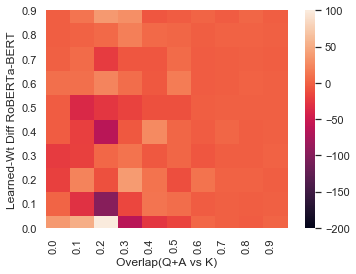}
  \caption{Difference between weights learned by the RoBERTa Weighted-Sum Model vs BERT-Weighted-Sum Model for the Normalized Overlap between knowledge and concatenated question-answer for all samples of PIQA validation set}
  \label{fig:diff}
\end{figure}

Here, both the learnt weights and the percentage overlap between the question and option versus knowledge is binned with an interval of 0.1. The figure \ref{fig:diff} shows the difference between the counts of samples for each weight and overlap bin between BERT-weighted-sum model and RoBERTa-weighted-sum model. From the figure, it can be seen that, the samples with percentage overlap between 0.2 to 0.3 which have been assigned lower weight in the region 0.0 to 0.1 are more in BERT weighted sum model than in RoBERTa weighted sum model.\\
On the other hand, the samples with percentage overlap between 0.2 to 0.3 which have been assigned weight in the region 0.1 to 0.2 and in region 0.2-0.3 are more in RoBERTA weighted sum model than BERT weighted sum model.

This shows the RoBERTa model is able to assign weights to proper knowledge sentences leading to improved question answering performance.

\subsection{Examples which need knowledge beyond commonsense knowledge}
There are some questions which need knowledge beyond common sense knowledge to answer. Following are 2 examples.
\fbox{\begin{minipage}{19em}
    \textbf{Goal:} how do you call the fuzz?\\ \textbf{(A) dial 911. }\\\textit{(B) dial fuzz under contacts.}
\end{minipage}}
\fbox{\begin{minipage}{19em}
    \textbf{Goal:} To fight Ivan Drago in Rocky for sega master system.?\\ \textit{(A) Drago isn't in this game because it was released before Rocky IV. }\\\textbf{(B) You have to defeat Apollo Creed and Clubber Lang first.}
\end{minipage}}

\end{document}